\documentclass[12pt]{l4dc23/l4dc2023}

\title[Predict-and-Critic]{Predict-and-Critic: Accelerated End-to-End Predictive Control for Cloud Computing through Reinforcement Learning}
\usepackage{times}

\usepackage{algcompatible}
\usepackage{amsmath, amssymb}
\usepackage{xcolor}
\usepackage{graphicx}
\usepackage{cite}
\usepackage{mathtools}
\usepackage{soul}
\usepackage{algorithm}
\usepackage{mathrsfs}
\usepackage{bbm}
\usepackage{todonotes}
\usepackage[autostyle]{csquotes}
\usepackage{hyperref}  
\hypersetup{
    colorlinks=true,
    citecolor=black,
    linkcolor=black,
}
\usepackage{multirow}
\usepackage[normalem]{ulem}
\useunder{\uline}{\ul}{}
\usepackage{color}
\definecolor{light}{rgb}{0.5, 0.5, 0.5}
\def\light#1{{\color{light}#1}}

\usepackage{amsmath,amsfonts,bm}

\def\1{\bm{1}}

\DeclareMathAlphabet{\mathsfit}{\encodingdefault}{\sfdefault}{m}{sl}
\SetMathAlphabet{\mathsfit}{bold}{\encodingdefault}{\sfdefault}{bx}{n}

\newcommand{\R}{\mathbb{R}}

\DeclareMathOperator*{\argmin}{arg\,min}

\newcommand{\E}{\mathop{\mathbb{E}}}
\newcommand{\D}{\mathcal{D}}
\newcommand{\cV}{\mathcal{V}}
\newcommand{\cL}{\mathcal{L}}

\newcommand{\w}{\omega}

\newcommand{\Z}{\mathbb{Z}}
\newcommand{\I}{\mathcal{I}}

\usepackage{xspace}
\newcommand{\cQPTL}{hcQPTL\xspace}
\newcommand{\cSPO}{hcSPO\xspace}
\newcommand{\TermInBracket}{(where the 'hc' represents hard constraint backprop)\xspace}
\newcommand{\Real}{AWS\xspace}

\author{%
 \Name{Kaustubh Sridhar\textsuperscript{\rm 1}\textsuperscript{\rm *}} \Email{ksridhar@seas.upenn.edu}\\
 \Name{Vikramank Singh\textsuperscript{\rm 2}} \Email{vkramas@amazon.com}\\
 \Name{Balakrishnan(Murali) Narayanaswamy\textsuperscript{\rm 2}} \Email{muralibn@amazon.com}\\
 \Name{Abishek Sankararaman\textsuperscript{\rm 2}} \Email{abisanka@amazon.com}\\
 \addr \textsuperscript{\rm 1}University of Pennsylvania, \textsuperscript{\rm 2}AWS AI Labs, \textsuperscript{\rm *}Work done while at Amazon
 \vspace{-2em}
}

\begin{document}

\maketitle

\begin{abstract}%
Cloud computing holds the promise of reduced costs through economies of scale. To realize this promise, cloud computing vendors typically solve sequential resource allocation problems, where customer workloads are packed on shared hardware. Virtual machines (VM) form the foundation of modern cloud computing as they help logically abstract user compute from shared physical infrastructure. Traditionally, VM packing problems are solved by predicting demand, followed by a Model Predictive Control (MPC) optimization over a future horizon. We introduce an approximate formulation of an industrial VM packing problem as an MILP with soft-constraints parameterized by the predictions. Recently, predict-and-optimize (PnO) was proposed for end-to-end training of prediction models by back-propagating the cost of decisions through the optimization problem. But, PnO is unable to scale to the large prediction horizons prevalent in cloud computing. To tackle this issue, we propose the Predict-and-Critic (PnC) framework that outperforms PnO with just a two-step horizon by leveraging reinforcement learning. PnC jointly trains a prediction model and a terminal Q function that approximates cost-to-go over a long horizon, by back-propagating the cost of decisions through the optimization problem \emph{and from the future}. The terminal Q function allows us to solve a much smaller two-step horizon optimization problem than the multi-step horizon necessary in PnO. We evaluate PnO and the PnC framework on two datasets, three workloads, and with disturbances not modeled in the optimization problem. We find that PnC significantly improves decision quality over PnO, even when the optimization problem is not a perfect representation of reality. We also find that hardening the soft constraints of the MILP and back-propagating through the constraints improves decision quality for both PnO and PnC.
\end{abstract}

\section{Introduction}\label{sec:intro}
Cloud computing holds the promise of reduced costs through economies of scale, as they allow vendors to consolidate demand \citep{marston2011cloud}. To realize this promise cloud computing vendors must solve a set of sequential resource allocation and control problems, where customer workloads are packed on shared hardware - making resource allocation and control a fundamental problem for cloud vendors \citep{zhang2016resource, gulati2011cloud}.

Virtual Machines (VMs) \citep{barham2003xen} are critical for both users and cloud service providers. They enable allocation of isolated compute resources to users on shared Physical Machines (PMs). Algorithms for packing VMs efficiently into PMs have been an important direction of research and practice \citep{bobroff2007dynamic,xiao2012dynamic}. These packing problems are stochastic, where the compute resource demands of instances vary with time; allow throttling, where a throttled instance is allocated fewer resources than its demand; include migrations; limitations on and delays in live migrations (\textit{i.e.} repacking) of instances; arrivals and departures; and unrestricted number of instances over time. We call this sto\underline{C}hastic, \underline{L}ive, thr\underline{O}ttled, and \underline{U}nrestricte\underline{D} packing problem as the CLOUD packing problem.

A class of approaches that has been highly successful in tackling packing problems is Model Predictive Control (MPC) \citep{mpc_1, mpc_1_refa, mpc_1_refc}. Typically, in MPC, a model is trained to predict the future demands over a time horizon. 
The packing decisions are then obtained by the solution of an optimization problem that minimizes cumulative costs over the future horizon using the predictions as ground truth. 
Instead of this two-stage approach, recently the end-to-end predict-and-optimize (PnO) paradigm \citep{spo_theory, qptl_decision_focused_learning, spo_full_and_relax} - where, the prediction model is trained to directly minimize cumulative costs over the horizon (measure by regret) - has been shown to be superior. 

MPC methods, both two-stage and end-to-end (PnO), exploit the advances into MILP solvers such as Xpress, Gurobi, and CPLEX by approximating the transition model by a set of mixed continuous-integer linear equations \citep{adhs, cahs}. However, this formulation is only approximate and does not model any environmental disturbances.
Further, these solvers become prohibitively slow with a large number of decision variables and constraints (endemic to the CLOUD problem); and both of which scale linearly with the prediction horizon. In contrast, purely model-free approaches can be trained end to end which improves performance with un-modeled problem aspects \citep{haarnoja2018soft} while requiring lower computational overhead at test time. However, they have difficulty with enforcing constraints \citep{ma2019combinatorial}, under-perform model-based methods \citep{fireplace, joshi2019efficient}, and are sample inefficient during training \citep{joshi2019efficient}.

This leads us to two questions, which we answer in the positive in this paper. 
Can we improve the computational complexity of PnO methods by integrating the strengths of model-free reinforcement learning (RL)? Can such a combined MPC + model free RL method perform well in the face of model errors?

To answer the above questions, we introduce an MILP formulation of the CLOUD problem which consists of soft-constraints that arise from the fact that allocations should not exceed future demand. Soft constraints can be violated but a violation incurs a penalty in the objective function \citep{hardening}. While the traditional approach is to add soft-constraints to the cost function, 
we can also 'harden' the soft constraints and backpropagate through the optimization problem's constraints. While backprop through constraints has not been explored before, the simple idea of 'hardening' soft constraints has been noted previously for 
improving solution quality by avoiding domain values at the extremes of the soft constraint \citep{hardening_2}. \textbf{Our Contributions:} 
\begin{enumerate}
    \item We propose predict-and-critic (PnC), a framework for integrating the strengths of model-free RL with sequential, multi-stage PnO algorithms. We utilize gradient-based feedback from both regret and the Q function to update a shared prediction model within deep deterministic policy gradient (DDPG) \citep{ddpg} framework. 
    This allows us to solve a much smaller MILP, while accounting for the future impact of decisions we make now.
    \item Since current PnO methods are restricted to problems where only the cost is parameterized by the prediction model, a secondary constribution of our work is in extending PnO methods -- QPTL \citep{qptl_decision_focused_learning} and SPO \citep{spo_full_and_relax}, to MILPs with hardened soft constraints parameterized by predictions.
    \item We demonstrate with both synthetic and industry datasets, three workloads and various migration delays that on the CLOUD packing problem, PnC outperforms PnO and that models trained with PnC are robust to stochasticity in migration delays. We also show that hardening the soft-constraints and backpropogating through the constraints improves solution quality in both PnO and PnC.
\end{enumerate}

\section{Related Work}\label{sec:related}

\noindent\textbf{On MPC for VM packing problems: }
In two-stage approaches to the packing problem \citep{mpc_1_refa, mpc_1_refc}, future demand is first forecast via minimizing a quantile error \citep{mqrnn} or mean squared error. This trained model is used at test time and the model's predictions are incorporated by the optimization problem to make a packing decision, typically assuming the predictions are accurate. Recent work on combining prediction and optimization has demonstrated that training a prediction model to directly minimize the optimization objective significantly improves decision quality \citep{spo_theory}. 

\noindent\textbf{On cost-parameterized PnO: }
PnO algorithms have been shown to surpass two-stage methods in a variety of combinatorial optimization problems primarily through the QPTL \citep{qptl_decision_focused_learning, qptl_with_cutting_planes_for_integral_solution} and SPO frameworks \citep{spo_theory, spo_full_and_relax}. Yet, they are unable to scale to large prediction horizons because of the increasing size of the optimization problem \citep{survey_sec7}. To improve scalability and decision quality, we  account for regret over the complete future (and yet only solve the optimization problem with the smallest possible horizon) in PnC. Further, QPTL and SPO methods cannot backpropagate through parameterized constraints. We propose extensions of QPTL and SPO that can do so and find that backprop through hardened soft-constraints improves decision quality in both PnO and PnC.

\noindent\textbf{On model-free augmentations for model-based methods: }
In the context of training predictive models for high quality decision making, the closest developments to PnC lie in robotics \citep{TD-MPC, buckman2018sample, Sikchi2020LearningOW, lowrey2018plan} where model-free RL has been shown to improve scalability and performance. Yet, robotics problems, with continuous (rather than discrete) optimization problems and fewer decision variables face fewer challenges with differentiability of optimization layers and optimization scale than the CLOUD packing problem.

\section{Background on Cost-Parameterized PnO} \label{sec:back}

\begin{figure*}[t]
    \centering
    \includegraphics[width=\linewidth]{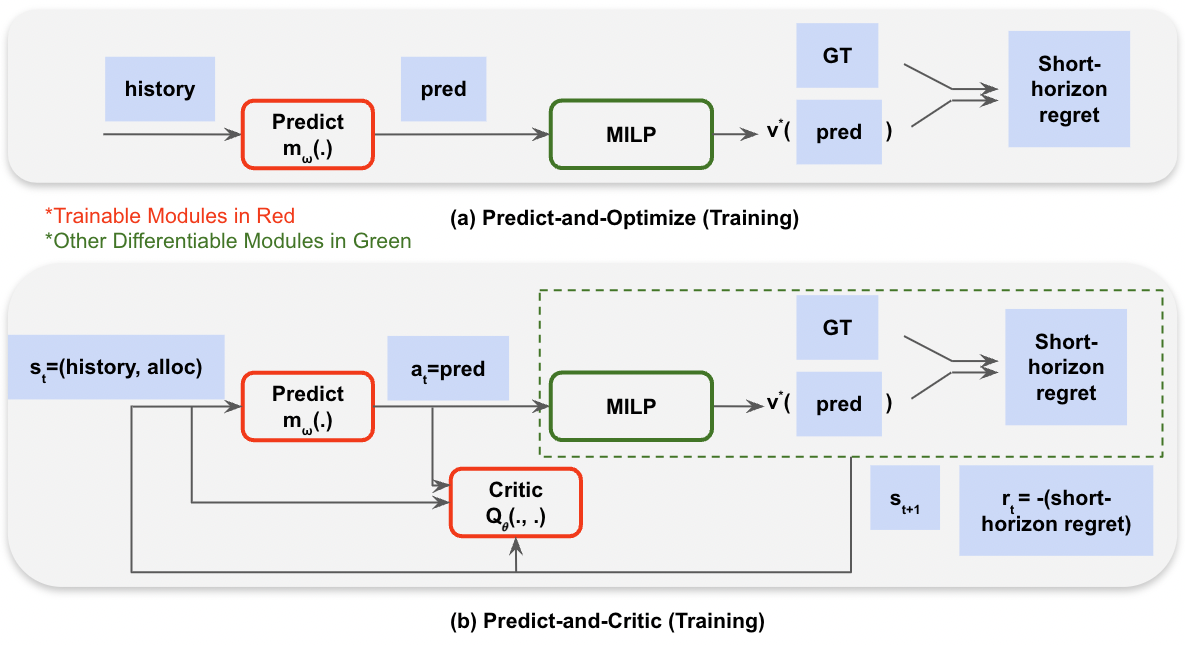}
    \includegraphics[width=0.475\linewidth]{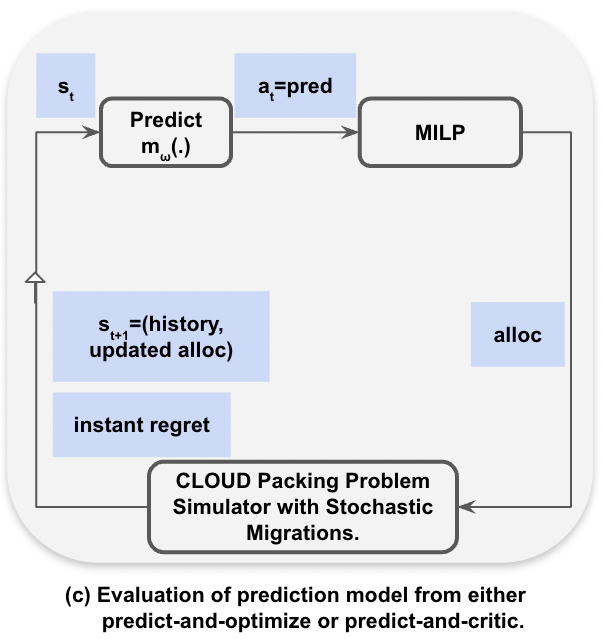}
    \caption{
    $(a, b)$ Predict-and-Optimize (Horizon $H>>2$) vs Predict-and-Critic (Horizon 2) for Training a Prediction Model. $(c)$ Evaluation of a prediction model and its corresponding MILP. A detailed version of this Figure can be found in the Appendix.
    }
    \label{fig:PC}
\end{figure*}

For cost-parameterized MILPs with decision variables $v \in \cV$ (formulated below),
{
\begin{align}
    v^*(\hat{y}) = \argmin_v \hat{y}^T v \;\text{ s.t. }\; Av \leq b, \; v_i \in \Z^{+} \; \forall \; i \in \I \label{obj_param_MILP}
\end{align}
}
with $regret(v^*(\hat{y}), y) = y^T v^*(\hat{y})$, the state-of-the-art methods, QPTL and SPO, are given below.\\
\noindent\textbf{QPTL} \citep{qptl_decision_focused_learning}: where an MILP is relaxed to a Quadratic Program using a small penalty term at training time as follows.
{
\begin{align}
    \argmin_v \hat{y}^T v + \nu ||v||_2^2 \;\text{ s.t. }\; Av \leq b
\end{align}
}
with $v \in conv(\cV)$ where $conv(\cV)$ denotes the convex hull of $\cV$. This enables computation of $\frac{dv^*(\hat{y})}{d\hat{y}}$ gradient by differentiating through the KKT conditions \citep{differentiable_qp_layer} as follows.
{\small 
\begin{align}
    \begin{bmatrix} \nabla_{v^*}^2 f(v^*, y) & A^T \\ diag(\lambda^*) A & diag(Av^* - b) \end{bmatrix} \begin{bmatrix} \frac{dv^*}{d\hat{y}} \\ \frac{d\lambda^*}{d\hat{y}} \end{bmatrix} = \begin{bmatrix} \frac{-d\nabla_{v^*} f(v^*, y)}{d\hat{y}} \\ 0 \end{bmatrix} \label{orig_system}
\end{align}
}
where $\nabla_{v^*}^2 f(v^*, y) = 2 \nu I$, $\frac{d\nabla_{v^*} f(v^*, y)}{d\hat{y}} = I$, and $\lambda^*(\hat{y})$ are the dual variables of the continuous QP relaxation. 
The gradient of regret with model parameters is obtained as follows.
{
\begin{align}
    \frac{d \; regret(v^*(\hat{y}), y)}{d\w} = \frac{d \; regret(v^*(\hat{y}), y)}{d\hat{y}} \; \frac{d\hat{y}}{d\w} = \left( \frac{d \; regret(v^*(\hat{y}), y)}{dv^*(\hat{y})} \; \frac{dv^*(\hat{y})}{d\hat{y}} \right) \; \frac{d\hat{y}}{d\w} \label{full_backprop}
\end{align}
}
where the first gradient is just the gradient of the objective with the decision variable, the second is from solving \eqref{orig_system}, and the third is obtained via backpropagation.

\textbf{SPO} \citep{spo_full_and_relax}: 
where 
a subgradient (rather than the exact gradient in QPTL) of a convex upper bound of the regret called SPO+ \citep{spo_theory} is used in the backward loop. This is given below.
{
\begin{align}
    \frac{d\,regret(v^*(\hat{y}), y)}{d\hat{y}} = v^*(y) - v^*(2\hat{y} - y)
\end{align}
}
Using the above, 
the gradient of regret with model parameters is again obtained via \eqref{full_backprop}. 
\section{Problem Formulation}

In this section, we present the CLOUD packing problem and its approximate MILP formulations. Of note, the MILP formulation does not model migration delays and is specific to a arrival and departure workload. But the simulator (depicted in Figure \ref{fig:PC}(c)) includes migration delays and various workloads to test prediction models on stochastic disturbances not included in the MILPs. 

\subsection{MILP with Soft Constraints (soft MILP)}
The objective of the CLOUD packing problem is to minimize the sum of costs incurred by new hosts (PMs) and from migration and throttling of VMs. We assume that there are a maximum of $N$ VMs at any time in a prediction horizon of $H$ steps. Each host has a maximum capacity $C$. 

The decision variables are (1) $\text{alloc}_{iht}$, the resources allocated to VM $i$ (if on host $h$) at time $t$; (2) $\text{used}_{ht}$ which is 1 if host $h$ is used at time $t$ and 0 otherwise; (3) $\text{placed}_{iht}$ which is 1 VM $i$ is assigned to host $h$ at time $t$ and 0 otherwise; (4) $\text{migr}_{iht}$ which is 1 if host $h$ is migrating VM $i$ at time $t$ and 0 otherwise.
The objective is given as follows.
{ 
\small 
\begin{align}
    \min \; \sum_{t=1}^{H} \bigg(& h_{\$} \sum_{h=1}^{N} \text{used}_{ht} + \frac{m_{\$}}{2} \sum_{h=1}^{N} \sum_{i=1}^{N}  \text{migr}_{iht} + th_{\$} \sum_{i=1}^N \text{throt}_{it} + \sum_{h=1}^{N} \sum_{i=1}^{N} \lambda_{iht} (\text{alloc}_{iht} - y_{i(t+1)} \, \text{placed}_{iht}) \bigg) \label{soft_obj_main}
\end{align}}
where the throttling fraction for a VM $i$ at time $t$ is given by $\text{throt}_{it} = {\big(y_{i(t+1)} - \sum_{h=1}^N \text{alloc}_{iht} \big)}/{C}$. The first three terms minimize the costs of new hosts, migrations and throttling where $h_{\$}, m_{\$}$ and $th_{\$}$ are the real costs of one new host, one migration and throttling one resource unit. The last term is a soft constraint that a VM $i$ at time $t$ can only be allocated an amount less than or equal to its upcoming true demand $\left(y_{i(t+1)}\right)$. 

The hard constraints of the problem are (i) the domains of the decision variables given by $alloc_{iht} \in \Z^{+}$, $\text{used}_{ht} \in \{0, 1\}$, $placed_{iht} \in \{0, 1\}$, $migr_{iht} \in \{0, 1\}$, and $throt_{it} \in \R$ (ii) the sum of resources allocated to VMs on each host should be less than or equal to the hosts capacity if it is in use $\left(\sum_{i=1}^N \text{alloc}_{iht} \leq C \; \text{used}_{ht} \right)$, (iii) each VM should only be on one host $\left(\sum_{h=1}^N \text{placed}_{iht} = 1\right)$, (iv) a VM should only be placed on a host if it was allocated non-zero resources on that host $\left(\text{placed}_{iht} \geq {\text{alloc}_{iht}}/{C}  \right)$, 
(v) the number of migrations involving any host is restricted to a constant value $\mathbf{M}$ $\left(\sum_{i=1}^N \text{migr}_{iht} \leq \mathbf{M} \right)$, (vi) a migration takes place only placement of a VM differs from $t$ to $t+1$ $\left( \lvert \text{placed}_{iht} - \text{placed}_{ih(t+1)} \rvert \leq \text{migr}_{iht} \right)$, and (vii) others that reduce the enumeration tree and enable faster computation with solvers. All of the above for all $i, h, t$. 

We denote the MILP with the objective \eqref{soft_obj_main} and hard constraints (i)-(vii) as the \textit{soft MILP}. Since we do not have the true demand in practice, we can replace the true demand with the prediction $\hat{y}_{i(t+1)}$ in the objective and abstract the problem as follows.
{\begin{align}
    v^*_{s}(\hat{y}_{:N,\,t+1:t+H}) = &\argmin \left( p^T v + \lambda^T( A(\hat{y}_{:N,\,t+1:t+H})\, v - b) + c(\hat{y}_{:N,\,t+1:t+H}) \right) \nonumber \\
    \text{ s.t. } \; &Gv \leq h, \; v_i \in \Z^{+} \; \forall \; i \in \I \label{v^*_s}
\end{align}}
where the subscript $s$ denotes \textit{soft}, $H$ denotes the prediction horizon and $\lambda$ denotes the Lagrangian multipliers. The notation $\hat{y}_{:N, \,t+1:t+H}$ represents the forecast of demands for $N$ VMs (or time series) from time $t+1$ to $t+H$, $y_{:N,\,t+1:t+H}$ represents true values of the demand for the $N$ VMs in that same horizon, $x_{:N,\,:t}$ denotes input features from the start until current time $t$ and consists of the history of true demands $y_{:N,\,:t}$ as well as covariates (such as the real time and date) from time $0$ to $t$ and for the prediction horizon $t+1$ to $t+H$.
Throughout the rest of this paper, we will refer to \eqref{v^*_s} as the \textit{soft MILP} (or \textit{Lagrangian relaxed} MILP).

\subsection{MILP with 'hardened' soft constraints (hard MILP)}
An alternative formulation is to simply add the soft constraints as hard constraints to the MILP. We call such a formulation as the \textit{hard MILP} and abstract it as follows.
{
\begin{align}
    v^*_{h}(\hat{y}_{:N,\,t+1:t+H}) &= \argmin \left(p^T v + c(\hat{y}_{:N,\,t+1:t+H})\right) \label{v^*_h} \\
    \text{ s.t. }\; A(\hat{y}_{:N,\,t+1:t+H})\, v &\leq b, \nonumber\\
    \; Gv &\leq h, \; v_i \in \Z^{+} \; \forall \; i \in \I \nonumber
\end{align}}
where the subscript $h$ denotes \textit{hard}. 


\section{Predict-and-Optimize with Back-propagation through Constraints}

For the \textit{soft MILP}, we solve it in the forward pass and compute the regret in \eqref{regret_s}. In the backward pass, the gradient of regret with the prediction model's parameters can be computed with standard QPTL and SPO as detailed in Section \ref{sec:back}.
{\begin{align}
    \hspace{-0.3cm} regret_{s}&(v^*(\hat{y}), y) = p^T  v^*_{s}(\hat{y}_{:N,t+1:t+H}) + \lambda^T A(y_{:N,t+1:t+H})v^*_{s}(\hat{y}_{:N,t+1:t+H}) + c(y_{:N,t+1:t+H}) \label{regret_s}
\end{align}}
For the \textit{hard MILP}, we compute the regret given below.
{\begin{align}
    regret_{h}(v^*(\hat{y}), y) = p^T v^*_{h}(\hat{y}_{:N,\,t+1:t+H}) + c(y_{:N,\,t+1:t+H}) \label{regret_h} 
\end{align}}
In the backward pass, we have to backprop through the constraints. To do so, we derive the gradient of regret in both the QPTL and SPO frameworks and name these extensions \cQPTL and \cSPO respectively \TermInBracket.
\subsection{Backpropagating through Constraints in QPTL (\cQPTL):}
We write the complete set of derivatives of the KKT conditions below. The following incorporates the terms omitted in both \citet{qptl_decision_focused_learning, differentiable_qp_layer} where constraints did not depend on the predictions. 
{\small 
\begin{align}
    \begin{bmatrix} \nabla_{v^*}^2 f(v^*, y) & \mathbb{A}^T \\ diag(\lambda^*) \mathbb{A} & diag(\mathbb{A}v^* - b) \end{bmatrix} \begin{bmatrix} \frac{dv^*}{d\hat{y}} \\ \frac{d\lambda^*}{d\hat{y}} \end{bmatrix}& = 
    \begin{bmatrix} \frac{-d\nabla_{v^*} f(v^*, y)}{d\hat{y}} - \frac{d\mathbb{A}^T \lambda^*}{d\hat{y}} \\ -diag(\lambda^*) \frac{d\mathbb{A} v^*}{d\hat{y}} \end{bmatrix} \label{complete_system} 
\end{align}}
where we still have $\nabla_{v^*}^2 f(v^*, y) = 2 \gamma I$ and $\lambda^*(\hat{y})$ are the  are the dual variables of the continuous QP relaxation but $\frac{d\nabla_{v^*} f(v^*, y)}{d\hat{y}} = 0$ and $\mathbb{A} = [A(\hat{y}), G]^T$. Further, specific to the MILP formulations of the CLOUD packing problem, we have $\frac{dG}{d\hat{y}}=0$. 
Hence, by solving \eqref{complete_system}, we obtain the gradient $\frac{dv^*(\hat{y})}{d\hat{y}}$.
Lastly, we can use \eqref{full_backprop} to find the gradient of regret with the prediction model's parameters for gradient descent.

\subsection{Backpropagating through Constraints in SPO (\cSPO):}
In SPO, since it is possible to use any oracle for solving the problem \citep{spo_full_and_relax}, we can utilize the solution of the \textit{hard MILP} in the backward pass as shown in \eqref{ha_spo}. Again, we can use \eqref{full_backprop} to find the gradient of regret with the prediction model's parameters for gradient descent. 
{\begin{align}
    \frac{d\,regret(v^*(\hat{y}), y)}{d\hat{y}}& = v^*_{h}(y_{:N,\,t+1:t+H}) - v^*_{h}(2\hat{y}_{:N,\,t+1:t+H} - y_{:N,\,t+1:t+H}) \label{ha_spo}
\end{align}}


{
\begin{algorithm}[t!]
    \small 
  \caption{\texttt{PnC} (DDPG Framework)}\label{alg:train_PC}
  \hspace*{\algorithmicindent} \textbf{Input: } $\;\;$Training sequence length $t_0$, total VMs/time series $n_{\text{total}}$, train dataset $D_{train} =$\\$\{(x_{0,\,:t_0}, y_{0,\,:t_0}), (x_{1,\,:t_0}, y_{1,\,:t_0}), ..., (x_{{n_{\text{total}}},\,:t_0}, y_{{n_{\text{total}}},\,:t_0})\}$, initial predict parameters $w$ and critic parameters $\theta$, initial empty allocation $alloc^*_0$.\\
  \hspace*{\algorithmicindent} \textbf{Output: } Trained prediction model $m_{\w}(.)$.
  \begin{algorithmic}[1]
    \STATE Initialize target parameters $\theta_{\text{targ}} \leftarrow \theta$, $\w_{\text{targ}} \leftarrow \w$, replay buffer $\D \leftarrow []$.
    \REPEAT
        \STATE Sample batch of $N$ time series from $D_{train}$
        \FOR{$t$ = $1, ..., t_0-1$}
            \STATE $\hat{y}_{:N,\,t+1} = m_{\w}(\left[x_{:N,\,:t}, \; alloc^*_{t-1} \right])$ \hfill  \light{// predict with policy/prediction model}
            \STATE Solve for $v^*(\hat{y}_{:N,\,t+1:t+2})$ \hfill \light{// \textit{hard} \eqref{v^*_h} or \textit{soft} \eqref{v^*_s} MILPs}
            \STATE Store the following in replay buffer $\D$:
            \begin{align*}
                \Bigg(
                \underbrace{\left[ x_{:N,\,:t}, alloc^*_{t-1} \right]}_{\textbf{state}},
                \underbrace{\hat{y}_{:N,\,t+1:t+2}}_{\textbf{action}}, 
                \underbrace{\left[ x_{:N,\,:t+1}, alloc^*_{t} \right]}_{\textbf{next state}}, 
                \underbrace{-regret(v^*(\hat{y}), y)}_{\textbf{reward}}, 
                \underbrace{\frac{d \; regret(v^*(\hat{y}), y)}{d\w}}_{\textbf{regret gradient}\text{ via PnO \eqref{ha_spo}, \eqref{complete_system}, \eqref{full_backprop}}}\Bigg)
            \end{align*}
            \vspace{-1.5em}
        \ENDFOR
        \IF{it's time to update}
            \FOR{however many updates}
                \STATE Sample batch $B$ from $\D$ and perform optimization update:
                \STATE $\;\;\;\;\;\theta \leftarrow \theta + \alpha_1 \nabla_{\theta} \cL_{1}(\theta, \w)$; $\;\;$ $\w \leftarrow \w + \alpha_2 \nabla_{\w} \cL_{2}(\w) + \alpha_1 \nabla_{\w} \cL_{1}(\theta, \w)$ \hfill \light{//via \eqref{both_w}, \eqref{both_theta}, \eqref{predict_w}}
                \STATE Update target networks: $\theta_{\text{targ}} \leftarrow \rho \theta_{\text{targ}} + (1-\rho)\theta$, $\;\;$ $\w_{\text{targ}} \leftarrow \rho \w_{\text{targ}} + (1-\rho) \w$
            \ENDFOR
        \ENDIF
    \UNTIL{\textbf{convergence}}
    \STATE \textbf{return} $m_{\w}(.)$
  \end{algorithmic}
\end{algorithm}
}

\section{Predict-and-Critic: Gradient-based Feedback from Regret and a Q-Function}


We can think of the predictor as an actor that takes in a state and outputs a continuous action. The MILP can be thought of as the environment which takes in the continuous action and outputs the next state and reward. The state is $s_t = \left[x_{:N, :t}, \text{alloc}^*_{t-1} \right]$, the action is given by $a_t  = m_{\w}(s_t) = \hat{y}_{:N, t+1:t+2}$, and the next state is $s_{t+1} = \left[x_{:N, :t+1}, \text{alloc}^*_{t} \right]$. The reward is the negative of regret and is denoted ${r_t}_{\w} = - \, regret(v^*(m_{\w}(s_t)), y_{:N, t+1:t+2})$.
With this intuition, we can train the actor (predictor) jointly with a stabilizing critic via the following losses in the forward pass of PnC (see Figure \ref{fig:PC}). These losses have been adapted to the PnC setting from DDPG \citep{ddpg}. 
{\begin{align} 
    \cL_{1}(\theta, \w) &= \E_{(s_t, a_t, s_{t+1}, {r_t}_{\w}) \sim \D} \Big( Q_{\theta}(s_t, a_t) - \big( {r_t}_{\w} + \gamma \; Q_{\theta_{\text{targ}}}(s_{t+1}, a_{t+1}) \big) \Big)^2\\
    \cL_{2}(\w) &= - \E_{s_t \sim \D} \left[ Q_{\theta}(x, m_{\w}(s_t)) \right]
\end{align}}
where $\D$ is a replay buffer that stores transitions $(s_t, a_t, s_{t+1}, {r_t}_\w, {d {r_t}_\w}/{d \w})$ as described in Algorithm \ref{alg:train_PC}.
The MILP solved by the environment is either the hard or soft MILP. In the backward pass the gradients are obtained for a batch $B$ as follows and updated as in Algorithm \ref{alg:train_PC}.
{\small 
\begin{align}
   \nabla_{\w} \cL_{1}(\theta, \w) = \frac{1}{|B|} \sum_{(s_t, a_t, s_{t+1}, {r_t}_{\w}, \frac{d {r_t}_{\w}}{d \w}) \in B} 2 \Bigg( Q_{\theta}(s_t, a_t) -\Big( {r_t}_{\w} + \gamma Q_{\theta_{\text{targ}}}(s_{t+1}, a_{t+1}) \Big) \Bigg) \Bigg( \frac{d Q_{\theta}}{{d \w}} -\frac{d {r_t}_{\w}}{{d \w}}\Bigg) \label{both_w}
\end{align}}
{\small
\begin{align}
    &\nabla_{\theta} \cL_{1}(\theta, \w) = \frac{1}{|B|} \sum_{(s_t, a_t, s_{t+1}, {r_t}_{\w}) \in B} 2 \Bigg( Q_{\theta}(s_t, a_t) - \Big( {r_t}_{\w} + \gamma \; Q_{\theta_{\text{targ}}}(s_{t+1}, a_{t+1}) \Big) \Bigg) \left( \frac{d \, Q_{\theta}(s_t, a_t)}{d \theta} \right) \label{both_theta}
\end{align}}
{\small
\begin{align}
    &\nabla_{\w} \cL_{2}(\w) = - \frac{1}{|B|} \sum_{s_t \in B} \frac{d Q(s_t, m_{\w}(s_t))}{d m_{\w}(s_t)}\, \frac{d m_{\w}(s_t)}{d \w}  \label{predict_w}
\end{align}}
We note that the term $\frac{d \, {r_t}_{\w}}{d \w}$ in \eqref{both_w} is computed from one of the PnO variants described before (LRSPO, LRQPTL, HASPO, H\cQPTL). We can also add exploration noise in Line 5 of Algorithm \ref{alg:train_PC} but observe that the gradients from regret (\textit{i.e.} $\nabla_{\w} \cL_1(\theta, \w)$), not usually present in the standard DDPG algorithm, provide a guided exploration noise by themselves. Finally, we can also set the gradient $\frac{d \, Q_{\theta}(s_t, a_t)}{d \w}$ in \eqref{both_w} to zero (\textit{i.e.} fixed Q function weights) to increase effectiveness of the gradient feedback from regret.

\section{Experimental Evaluation}
To evaluate a prediction model and its corresponding MILP, at every timestep, the model makes a prediction over a horizon which is given to the MILP to generate an optimal allocation for the current time. This allocation is fed to and updated by the simulator with unmodeled disturbances (such as migration delays and changes in workload). The simulator also outputs input features for the next timestep and a regret value (of new hosts, migrations and throttling \Real cost). We accumulate these regrets to compare methods. 

\textbf{Synthetic data: } 
We create three synthetic datasets and two industrial datasets. The synthetic datasets consist of VMs with sinusoidally varying demands at different frequencies: low, high and mixed. In the low frequency sinusoidal dataset, we simulate VM demand with increasing time $t$ as $55+25sin(it/100)$ for $i=1, ..., N$; in high frequency as $55+25sin(100it)$ for $i=1, ..., N$; and in mixed as $55+25sin(it)$ for $i=1, ..., N$. The timegap between two consecutive timesteps is 12 hours. We also set $N=10$. This means that while there can be an unbounded number of VMs that can arrive and depart, there can only be 10 active VMs at any point in time.

\textbf{Industrial data: } 
In the industrial datasets, we have demands of compute tasks obtained at 1 minute intervals from Amazon Web Services. We set $N=10$ in one such dataset and $N=100$ in another dataset. The sinusoidal datasets are named Mixed10, High Frequency10, Low Frequency10; the industry datasets are named \Real10, and \Real100. 

\textbf{Complexity and Scale: } 
\textit{The number of decision variables scales as $O(N^2 H)$.} With a horizon of 5, and $N=10$ itself, each MILPs has over $2000$ decision variables and $7000$ constraints. We restrict our training time $t_0$ to 50 timesteps (25 days in the sinusoidal data and 50 min in industrial data) for each of the $n_{\text{total}}$ time series. 
In an epoch, we select a batch of $N$ time series from $n_{\text{total}}$ time series'. For each batch, at each training timestep (out of 50), we have a prediction of size $(N, H)$ and a corresponding MILP. For training, we have to solve 50 MILPs in each batch of each epoch. This already exceeds the $48$ VMs limit per epoch in \citet{spo_full_and_relax}.
Moreover, with $N=100$, the decision variables exceed $200,000$ and constraints $700,000$. To repeatedly solve this problem for $t_0$ timesteps in each batch of each epoch of training, we use a sparsification heuristic where we only pack newly arriving and departing VMs and migrate 10 other VMs at each point in time.
We simulate all methods at test time for 25 timesteps on $N$ VMs (300 hours of sinusoidal data and 25 min of industrial data). 

\textbf{Workloads: }
A particular workload is a description of the arrival and departure pattern of VMs. We use Burst (B) when all VMs arrive at the start, Gradual (G) where VMs arrive gradually and Cyclic (C) where VMs arrive and depart cyclically and staggered. We set gradual arrival as one VM every 2 timesteps (1 day in sinusoidal data and 2 min in industrial data). We perform staggered cyclic arrival at a period of 10 timesteps (5 days in sinusoidal data and 10 min in industrial data). 

\textbf{Migration Delays: }
We also simulate migration delays of 2 timesteps (24 hr for sinusoidal data and 2 min for industrial data) and 5 timesteps (60 hr for sinusoidal data and 5 min for industrial data). During a migration delay, allocations of VMs do not change from their pre migration value. 

\textbf{Model, Warmstarting and Solver: }
We utilize a multi-quantile recurrent neural network \citep{mqrnn} trained with a quantile (median) loss to predict future demand in the two stage approach. We use 5 hidden recurrent layers and 100 neurons in each layer. We warmstart the model like in \citet{spo_full_and_relax} for PnO and PnC by using the weights of the two stage median-predicting model. We train all models with a learning rate of $0.001$, an Adam optimizer and for 10,000 epochs. This includes a training time of $10000*50*(\text{time to solve the corresponding MILP})$. We set $\alpha_1=0.05$, $\alpha_2=0.95$, and $\rho=0.95$ in Algorithm \ref{alg:train_PC}. We obtained these values with a grid search over $\alpha_1, \alpha_2$ in $[0.025, 0.975]$. We also perfomed a grid search over learning rate in $[10^{-4}, 10^{-1}]$. We use the Xpress solver for solving MILPs. We answer several research questions below.


\textbf{\texttt{(RQ1)} Short-Horizon PnC vs Long-Horizon PnO vs Long-Horizon Two-stage MPC: }
We analyze how PnC with the smallest instantiation of the optimization problem (\textit{i.e.,} $H=2$) compares with longer horizon PnO. We notice in Table \ref{tab:rq2} that PnC outperforms PnO and Two-stage MPC on all datasets. We also notice consistent improvements across all four regret gradient methods (SPO, \cSPO, QPTL, \cQPTL) on the Mixed10, HF10 and \Real10 datasets. We attribute this to the critic's gradient-feedback from beyond the prediction horizon. 

Further, even though PnO is not as good as PnC, as we expect, PnO with a larger horizon of $5$ performs (in general) better than with a smaller horizon of $4$. Also, as expected, PnO outperforms two-stage model predictive control. This too can be observed from Table \ref{tab:rq2}.

\begin{table}[t]
\centering 
\scriptsize 
\begin{tabular}{c|c|c|c|c||c}
\hline \hline 
\multicolumn{2}{c|}{Methods} & \multicolumn{4}{c}{Regret on various datasets}  \\ \cline{3-6} 
\multicolumn{2}{c|}{} &  Mixed10 & HF10 & LF10 & \Real10 \\ \hline \hline 
Heuristics & First Fit & 616.53 & 618.05 & 602.36 & 401.16  \\
 & Best Fit  & 602.83 & {611} & \textbf{384.29} & 229.93 \\ \hline 
Two-stage MPC & soft MILP  & 1044.19  & 1059.13  & 1041.18   & 554.94   \\ 
 ($H=4$) & hard MILP  & 955.19  & 920.76 & 941.01  & 546.71    \\ \hline 
PnO & SPO  & 638.74  & 663.69  & 637.17  & 165.42   \\ 
 ($H=4$) & \cSPO  & 624.58  & 645.55  & 549.23 & 157.2   \\ 
 & QPTL  & 1024.89  & 1085.85  & 1095.18  & 660.05   \\ 
 & \cQPTL  & 605.02  & 1014.4  & 919.19    & 627.14   \\ \hline 
Two-stage MPC & soft MILP  & 1080.31  & 1073.48  & 1003.66 & 563.5   \\
 ($H=5$) & hard MILP  & 932.14  & 956.12 & 944.58  & 556.27 \\ \hline 
PnO& SPO  & 638.46  & 642.35  & 601.57     & 216.59 \\
($H=5$) & \cSPO  & {621.47}  & {597.53}  & {533.32}   & {152.7} \\ 
 & QPTL  & 951.93  & 1063.67  & 1087.92 & 568.73  \\
 & \cQPTL  & {600.81}  & {631.97}  & {1075.65} & 553.37  \\ \hline
PnC \textbf{(ours)} & SPO  & 639.16  & {644.16}  & 634.39 & \textbf{135.51} \\
($H=2$) & \cSPO  & \textbf{572.49}  & \textbf{535.27}  & 589.48   & \textbf{135.51}   \\
 & QPTL  & 611.27  & 1003.58  & {597.5}  & \textbf{135.51} \\
 & \cQPTL  & {591.27}  & 998.35 
 & 597.5  & \textbf{135.51} \\
\hline \hline 
\end{tabular}
\caption{\texttt{(RQ1)} PnC with $H=2$ vs PnO with $H=4, 5$ vs Two-stage MPC with $H=4, 5$. \texttt{(RQ2)} PnC vs Heuristics. The lowest regret for each column is highlighted in \textbf{bold}. 
}
\label{tab:rq2}
\vspace{-1em}
\end{table}

\textbf{\texttt{(RQ2)} Comparison with Online Heuristics: }
In Table \ref{tab:rq2} on the sinusoidal dataset, we notice that with high frequency and mixed datasets, PnC substantially improves upon PnO and the online heuristics of first fit and best fit. But on the low frequency, we notice that best fit outperforms all methods. With low frequency of changes in demands, throttling (the only objective term with the prediction in it) costs are minimal to zero. Since the heuristics choose a packing and do not migrate, they avoid migration costs unlike others and with comparable hosts costs, turn out to be the best. 

\textbf{\texttt{(RQ3)} Effects of Scale with \Real100: }
On the \Real100 dataset, we still observe that PnC methods outperform PnO by a large margin!

\textbf{\texttt{(RQ4)} Unmodeled Workloads and Migration Delays: }
We compare the robustness of PnO and PnC to migration delays and to the gradual and cyclic workloads that were not encoded into the MILP. We see in Table \ref{tab:rq4_workload} that PnC continues to beat PnO on unmodeled workloads (gradual and cyclic). We also notice an increase in regret from Burst to Gradual and then a decrease to Cyclic. The increase from Burst to Gradual occurs because of the later arrivals that are not present in the start which induce throttling. The overall decrease to Cyclic takes place because fewer hosts are required with staggered arrivals and departures continuously happening. Also, in Table \ref{tab:rq4_workload}, we observe that under migration delays of 2 min, PnC is still the best choice, but with increasing delay to 5 min, PnO with \cSPO is more performant.

\begin{table}[t!]
\centering
\begin{minipage}{.4\textwidth}
    \centering
    \scriptsize 
    \begin{tabular}{c|c|c}
    \hline \hline 
    
    \multicolumn{2}{c|}{Methods} & Regret on \\ \multicolumn{2}{c|}{} & \Real100 \\ \multicolumn{2}{c|}{} & dataset \\ \hline \hline 
    PnO & SPO & 1957.38 \\ 
    ($H=5$) & \cSPO & 1796.55 \\ 
    & QPTL & 1781.80 \\ 
    & \cQPTL & 3921.01 \\ \hline 
    PnC \textbf{(ours)} & SPO & \textbf{1709.05} \\ 
    ($H=2$) & \cSPO &  \textbf{1709.05} \\ 
    & QPTL & {1710.05} \\ 
    & \cQPTL & {1710.05} \\
    \hline \hline 
    \end{tabular}
    \caption{\texttt{(RQ3)} Effects of Scale. 
    }
    \label{tab:rq6}
\end{minipage} \hfill 
\begin{minipage}{.59\textwidth}
    \centering
    \scriptsize 
    \begin{tabular}{c|c|c|c|c|c|c}
    \hline \hline 
    \multicolumn{2}{c|}{Methods} & \multicolumn{5}{c}{Regrets on \Real10 dataset}  \\ \cline{3-7} 
    \multicolumn{2}{c|}{} &  \multicolumn{3}{c|}{Workloads} & \multicolumn{2}{c}{Migr. Delays} \\ \cline{3-7} 
    \multicolumn{2}{c|}{} &  Burst & Gradual & Cyclic & 2 min & 5 min  \\ \hline \hline

    PnO & SPO  & 216.59  & 236.59  & 201.03 & 222.59 & 214.95\\
    $H=5$ & \cSPO  & 152.7  & 211.19  & 198.62  & 214.06 & \textbf{167.33}\\
     & QPTL  & 568.73  & 568.75  & 493.39 & 520.64 & 481.56\\
     & \cQPTL  & 553.37  & 553.62  & 212.31 & 585.54 & 588.58\\ \hline 
    PnC \textbf{(ours)} & SPO  & \textbf{135.51}  & \textbf{135.72}  & \textbf{108.56}  & \textbf{138.51} & {173.51}\\
    $H=2$ & \cSPO  & \textbf{135.51}  & \textbf{135.72}  & \textbf{108.56} & \textbf{138.51} & {173.51}\\
     & QPTL  & \textbf{135.51}  & \textbf{135.72}  & \textbf{108.56} & \textbf{138.51} & 320.84\\
     & \cQPTL  & \textbf{135.51}  & \textbf{135.72}  & \textbf{108.56} & \textbf{138.51} & 369.95\\

    \hline \hline 
    \end{tabular}
    \caption{
    \texttt{(RQ4)} Workloads and Migration Delays. 
    }
    \label{tab:rq4_workload}
\end{minipage}
\end{table}

\textbf{(RQ5): Backpropagating through hardened 'soft' constraints vs soft constraints in the cost function: }
The final research question is on the comparison between backpropagating gradient of regret via the constraints ('hardened' soft constraints) or the cost function (soft constraints). In Table \ref{tab:rq2}, with results on the Burst workload, we notice that \cSPO and \cQPTL beat SPO and QPTL respectively on both types of data. Finally, we ask if PnC is impacted by the choice of method to obtain gradients of regret. On the synthetic dataset with the burst workload, we notice that indeed \cSPO performs better than SPO and \cQPTL outperforms QPTL. But, on the \Real datasets, we find that all four methods perform similarly with no clear winner. This can be observed in Tables \ref{tab:rq2}, \ref{tab:rq6} and \ref{tab:rq4_workload}. We suspect this equivalence between methods is due to the stabilizing nature of the gradient from the Q function. Additional results are in the Appendix.

\vspace{-1em}
\section{Conclusions and Future Work}\label{sec:conc}
In this work, we proposed the Predict-and-Critic (PnC) framework which leverages gradient-based feedback from both regret (like Predict-and-Optimize) and a terminal Q-function to improve upon Predict-and-Optimize (PnO) in long-horizon cloud computing problems.
We evaluated our methods on 2 types of datasets, 3 workloads and migration delays to demonstrate that PnC outperforms PnO even in the face of disturbances not included in the optimization problem. We also showed that back-propagating through hardened soft constraints (prevalent in cloud computing) rather than soft-constraints in the cost improves performance in both PnC and PnO. In future work, we will include the simulator in the training loop of PnC and extend our methods to water conservation, epidemiology and robotics.

\appendix


\bibliography{aaai23}

\newpage 
\section*{Appendix: Detailed Figure and Additional Experimental Evaluations}

\begin{figure}[H]
    \centering
    \includegraphics[width=\linewidth]{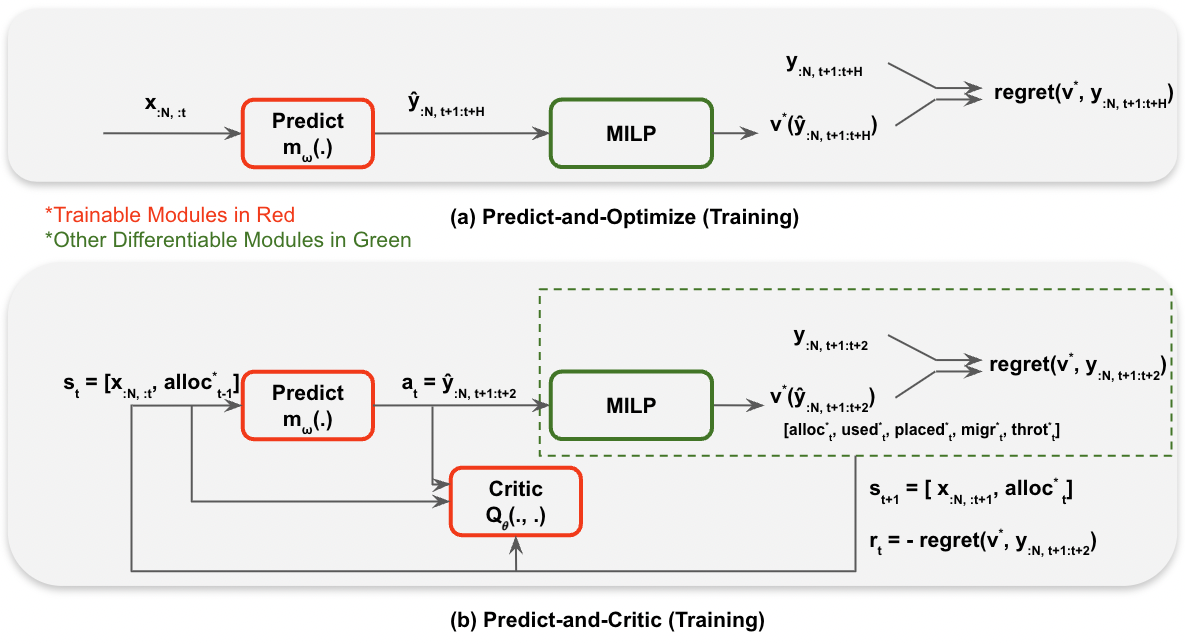}
    \includegraphics[width=0.475\linewidth]{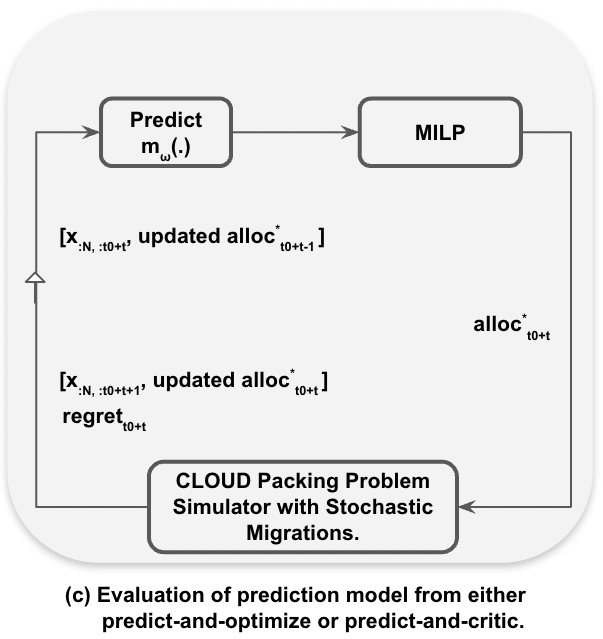}
    \caption{
    $(a, b)$ Predict-and-Optimize (Horizon $H>>2$) vs Predict-and-Critic (Horizon 2) for Training a Prediction Model. $(c)$ Evaluation of a prediction model and its corresponding MILP. 
    }
    \label{fig:PC_detailed}
\end{figure}

We tabulate the regrets with perfect forecasts (from an oracle) for a short simulation time of 5 timesteps in Table \ref{tab:perfect}. In Table \ref{tab:perfect}, we observe that the hard MILP attains lower regret than the soft MILP. This is because allocations chosen by the soft MILP are near the lower bound of 1, increasing throttling. This aligns with the extreme domain values observed with soft constraints in \citet{hardening_2}.

\begin{table}[H]
\centering 
\small 
\begin{tabular}{cc|cccc}
\hline \hline 
\multicolumn{2}{c|}{\multirow{2}{*}{Methods}}                                                                                      & \multicolumn{4}{c}{Datasets}                                                                               \\ \cline{3-6} 
\multicolumn{2}{c|}{}                                                                                                              & \multicolumn{1}{c|}{Mixed10}   & \multicolumn{1}{c|}{HF10} & \multicolumn{1}{c|}{LF10}  & Real10     \\ \hline \hline 
\multicolumn{2}{c|}{Hard MILP}                                                                                            & \multicolumn{1}{c|}{46.81}          & \multicolumn{1}{c|}{46.99}        & \multicolumn{1}{c|}{44.16}          &     8.50      \\ \hline
\multicolumn{2}{c|}{Soft MILP}                                                                                            & \multicolumn{1}{c|}{107.12}          & \multicolumn{1}{c|}{105.45}        & \multicolumn{1}{c|}{108.83}          &     26.56      \\ \hline \hline 
\end{tabular}
\caption{Comparison of optimal test regrets of the primary, {hard} and {soft MILP} with perfect forecasts for a short simulation of 5 timesteps. 
}
\label{tab:perfect}
\end{table}

Further, we tabulate the complete results across horizons and workloads for Heuristics, PnO and PnC on the Real10 dataset in Table \ref{tab:aws10}, and on various sinusoidal datasets (Mixed10, High Frequency10, Low Frequency10) in Table \ref{tab:data10}. We also tabulate the results across migration delays on the Mixed10 and Real10 datasets under a burst workload in Table \ref{tab:all_migr}. We note that we grid search over $\gamma$ values in $\{0.9, 0.95, 0.99\}$. 

\begin{table}[H]
\centering 
\scriptsize
\begin{tabular}{c|c|ccc}
\hline \hline 
\multicolumn{2}{c|}{\multirow{3}{*}{Methods}}                                                                                       & \multicolumn{3}{c}{Datasets / Workloads}                                                                                                                                                    \\ \cline{3-5} 
\multicolumn{2}{c|}{}                                                                                                               & \multicolumn{3}{c}{Real10}                   \\ \cline{3-5} 
\multicolumn{2}{c|}{}                                                                                                               & B & G & C\\ \hline \hline 
\multicolumn{1}{c|}{\multirow{2}{*}{Heuristics}}                                                                        & First Fit &       401.16                & 402  & 353.16  \\
\multicolumn{1}{c|}{}                                                                                                   & Best Fit  &         229.93              & 229.93  & 121.76  \\ \hline \hline 

PnO & Two-stage (soft) & 581.78  & 581.78  & 572.19   \\
($H=2$) & LRSPO & 173.51  & 173.51  & 149.02   \\
& HASPO & 167.09  & 167.09  & 145.75   \\
& LRQPTL & 669.69  & 569.69  & 545.2   \\
& HCQPTL & 592.66  & 592.66  & 571.41   \\ \hline
PnO & Two-stage (soft)& 556.71  & 556.71  & 570.09   \\
($H=3$) & LRSPO & 169.65  & 205.65  & 166.81   \\
& HASPO & 157.38  & 202.33  & 178.44   \\
& LRQPTL & 660.8  & 145.65  & 432.9   \\
& HCQPTL & 634.15  & 634.15  & 635.41   \\ \hline
PnO & Two-stage (soft) & 554.94  & 554.94  & 556.05   \\
($H=4$) & LRSPO& 165.42  & 207.63  & 157.08   \\
&HASPO & 157.2  & 208.43  & 191.2   \\
&LRQPTL & 660.05  & 156.19  & 480.71   \\
&HCQPTL & 627.14  & 627.14  & 607.17   \\ \hline
PnO & Two-stage (soft)& 563.5  & 533.75  & 574.92   \\
($H=5$) &LRSPO & 216.59  & 236.59  & 201.03   \\
&HASPO & 152.7  & 211.19  & 198.62   \\
&LRQPTL & 568.73  & 568.75  & 493.39   \\
&HAQPTL & 553.37  & 553.62  & 212.31   \\ \hline
PnC &LRSPO & \textbf{135.51}  & \textbf{135.72}  & \textbf{108.56}   \\
($H=2$) &HASPO & \textbf{135.51}  & \textbf{135.72}  & \textbf{108.56}   \\
&LRQPTL & \textbf{135.51}  & \textbf{135.72}  & \textbf{108.56}   \\
&HAQPTL & \textbf{135.51}  & \textbf{135.72}  & \textbf{108.56}   \\

\hline \hline 
\end{tabular}
\caption{Comparison of  Heuristics, PnO ($H=2, 3, 4, 5$) with PnC ($H=2$) on the Real10 dataset with Burst (B), Gradual (G), and Cyclic (C) Workloads. The \textbf{bold} value represents the least test regret in each column.  
}
\label{tab:aws10}
\end{table}

\begin{table*}[]
\centering 
\scriptsize
\begin{tabular}{c|c|ccc|ccc|ccc}
\hline \hline 
\multicolumn{2}{c|}{\multirow{3}{*}{Methods}}                                                                                       & \multicolumn{9}{c}{Datasets / Workloads}                                                                                                                                                    \\ \cline{3-11} 
\multicolumn{2}{c|}{}                                                                                                               & \multicolumn{3}{c|}{Mixed10}                       & \multicolumn{3}{c|}{High Freq10}                       & \multicolumn{3}{c}{Low Freq10}                      \\ \cline{3-11} 
\multicolumn{2}{c|}{}                                                                                                               & \multicolumn{1}{c}{B} & G & \multicolumn{1}{c|}{C} & \multicolumn{1}{c}{B} & G & \multicolumn{1}{c|}{C} & \multicolumn{1}{c}{B} & G & \multicolumn{1}{c}{C}\\ \hline \hline 
\multicolumn{1}{c|}{\multirow{2}{*}{Heuristics}}                                                                        & First Fit &       616.53                & 616.54  & \multicolumn{1}{l|}{611.8}  &     618.05                  &  618.6 &  \multicolumn{1}{l|}{613.14}  &           602.36            & 602.1  & \multicolumn{1}{l}{696.58} \\
\multicolumn{1}{c|}{}                                                                                                   & Best Fit  &         602.83              & 602.6  & \multicolumn{1}{l|}{632.81}  & 611                    &  616.74 & 618.02  &          \textbf{384.29}             & \textbf{386}  & \multicolumn{1}{l}{546.97} \\ \hline \hline

PnO & Two-stage  (soft)  & 887.3  & 887.3  & 766.37  & 797.34  & 797.34  & 684.8  & 883.86  & 883.86  & 770.25   \\
($H=2$) & LRSPO  & 739.16  & 739.16  & 719.85  & 744.16  & 744.16  & 732.77  & 734.39  & 734.39  & 721.33   \\
 & HASPO  & 718.09  & 718.09  & 684.9  & 707.51  & 707.51  & 690.75  & 787.3  & 787.45  & 631.28   \\
 & LRQPTL  & 1064.69  & 1064.69  & 944.59  & 1098.62  & 1098.62  & 986.95  & 1050.81  & 1050.81  & 936.65   \\
 & HCQPTL  & 920.56  & 920.56  & 830.69  & 987.25  & 987.25  & 880.72  & 1028.8  & 1028.8  & 940.56   \\ \hline
PnO & Two-stage  (soft)  & 1056.12  & 1056.12  & 933.56  & 1030.84  & 1030.84  & 981.06  & 1023.74  & 1023.74  & 905.27   \\
($H=3$) & LRSPO  & 639.34  & 639.34  & 620.04  & 643.51  & 643.51  & 532.12  & 635.82  & 635.82  & 522.77   \\
 & HASPO  & 635.78  & 635.78  & 532.27  & 580.0  & 546.79  & 526.68  & 682.6  & 682.6  & 569.18   \\
 & LRQPTL  & 1034.54  & 1034.54  & 929.49  & 1086.15  & 1086.15  & 947.65  & 1062.28  & 1062.28  & 961.39   \\
 & HCQPTL  & 837.2  & 1028.97  & 637.09  & 610.55  & 610.55  & 599.58  & 979.89  & 979.89  & 491.8   \\ \hline
PnO & Two-stage  (soft)  & 1044.19  & 1044.19  & 951.19  & 1059.13  & 1059.13  & 920.77  & 1041.18  & 1041.18  & 941.01   \\
($H=4$) & LRSPO  & 638.74  & 640.31  & 520.87  & 663.69  & 663.69  & 531.99  & 637.17  & 637.17  & 524.11   \\
 & HASPO  & 624.58  & 624.58  & 519.3  & 645.55  & 645.55  & 509.5  & 549.23  & 549.23  & 480.6   \\
 & LRQPTL  & 1024.89  & 610.71  & 690.32  & 1085.85  & 1085.85  & 986.77  & 1095.18  & 620.0  & 942.06   \\
 & HCQPTL  & 605.02  & 605.02  & 637.03  & 1014.4  & 1014.4  & 936.95  & 919.19  & 981.31  & 628.76   \\ \hline
PnO & Two-stage  (soft)  & 1080.31  & 1080.31  & 932.15  & 1073.48  & 1073.48  & 956.12  & 1003.66  & 1003.66  & 944.58   \\
($H=5$) & LRSPO  & 638.46  & 638.46  & 519.16  & 642.35  & 642.35  & 530.96  & 601.57  & 638.46  & 525.41   \\
 & HASPO  & 621.47  & 601.57  & 516.27  & 597.53  & 597.53  & 488.65  & 533.32  & 617.44  & 500.04   \\
 & LRQPTL  & 951.93  & 610.57  & 489.19  & 1063.67  & 614.46  & 958.4  & 1087.92  & 1087.92  & 960.61   \\
 & HCQPTL  & 600.81  & 604.81  & 624.18  & 631.97  & 631.97  & 634.93  & 1075.65  & 1075.65  & 971.93   \\ \hline
PnC & LRSPO  & 639.16  & 639.16  & 519.85  & 644.16  & 644.16  & 532.77  & 634.39  & 634.39  & 521.33   \\
($H=2$) & HASPO  & \textbf{572.49}  & \textbf{572.49}  & \textbf{465.55}  & \textbf{535.27}  & \textbf{535.27}  & \textbf{441.61}  & 589.48  & 589.48  & \textbf{482.36}   \\
 & LRQPTL  & 602.27  & 602.27  & 480.88  & 1003.58  & 1003.58  & 891.64  & 597.5  & 597.5  & \textbf{482.36}   \\
 & HCQPTL  & 611.27  & 611.27  & 489.88  & 998.35  & 998.35  & 888.07  & 597.5  & 597.5  & \textbf{482.36}   \\

\hline \hline 
\end{tabular}
\caption{Comparison of Heuristics, PnO ($H=2, 3, 4, 5$) with PnC ($H=2$) on the Mixed10, High Freq10, and Low Freq10 datasets with Burst (B), Gradual (G), and Cyclic (C) Workloads. The \textbf{bold} values represent the least test regret in each column. 
}
\label{tab:data10}
\end{table*}

\begin{table}[]
\centering 
\scriptsize
\begin{tabular}{c|c|cc|cc}
\hline \hline 
\multicolumn{2}{c|}{\multirow{3}{*}{Methods}}  & \multicolumn{3}{c}{Datasets / Migration Delays}  \\ \cline{3-6} 
\multicolumn{2}{c|}{}                                       &  \multicolumn{2}{c|}{High Freq10}  & \multicolumn{2}{c}{Real10} \\\cline{3-6}
\multicolumn{2}{c|}{}                                       & 24h & 60h & 2m & 5m\\ \hline \hline 
\multicolumn{1}{c|}{\multirow{2}{*}{Heuristics}}            & First Fit & 63.05 &  64.6 &  401.16 & 401.17\\
\multicolumn{1}{c|}{}                                       & Best Fit  & 110.74 &  110.75 & 145.23 & 149.42\\ \hline \hline

PnO & Two-stage (soft) &  789.71 &  673.66 &   370.81 &  370.81 \\
($H=2$) & LRSPO &  644.16 &  644.16 &   173.51 &  173.51 \\
 & HASPO &  598.06 &  598.94 &   167.01 &  \textbf{107.29} \\
 & LRQPTL &  1100.57 &  923.63 &   569.79 &  483.95 \\
 & HCQPTL &  985.14 &  793.36 &   591.31 &  563.93 \\ \hline
PnO & Two-stage (soft) &  1027.12 &  817.75 &   400.69 &  400.69 \\
($H=3$) & LRSPO &  643.51 &  643.51 &   205.42 &  200.69 \\
 & HASPO &  616.61 &  641.93 &   201.92 &  169.42 \\
 & LRQPTL &  1078.15 &  954.93 &   534.47 &  598.53 \\
 & HCQPTL &  691.89 &  681.15 &   638.38 &  510.99 \\ \hline
PnO & Two-stage (soft) &  1049.31 &  819.56 &   296.42 &  296.42 \\
($H=4$) & LRSPO &  663.69 &  700.58 &   203.72 &  224.42 \\
 & HASPO &  643.56 &  {592.21} &   204.19 &  124.99 \\
 & LRQPTL &  1084.08 &  1014.31 &   548.15 &  630.34 \\
 & HCQPTL &  1007.45 &  914.82 &   617.79 &  604.89 \\ \hline
PnO & Two-stage (soft) &  1055.71 &  817.86 &   281.95 &  281.95 \\
($H=5$) & LRSPO &  642.35 &  642.35 &   222.59 &  214.95 \\
 & HASPO &  {589.97} &  640.5 &   214.06 &  {167.33} \\
 & LRQPTL &  1073.76 &  1091.9 &   520.64 &  481.56 \\
 & HCQPTL &  717.59 &  684.77 &   585.54 &  588.58 \\ \hline
PnC & LRSPO &  644.16 &  644.16 &   \textbf{138.51} &  173.51 \\
($H=2$) & HASPO &  \textbf{535.27} &  \textbf{535.27} &   \textbf{138.51} &  173.51 \\
& LRQPTL &  973.42 &  1038.16 &   \textbf{138.51} &  320.84 \\
 & HCQPTL &  1052.57 &  942.72 &   \textbf{138.51} &  369.95 \\

\hline \hline 
\end{tabular}
\caption{Comparison, under migration delays, of PnO ($H=2, 3, 4, 5$) with PnC ($H=2$) on the Mixed10 and Real10 datasets with burst workloads. The \textbf{bold} values represent the least test regret in each column. }
\label{tab:all_migr}
\end{table}

\end{document}